\documentclass[a4paper]{llncs}

\usepackage{nomencl}
\usepackage{amssymb}
\usepackage{graphicx}
\usepackage{soul,color}
\usepackage{array, booktabs}
\usepackage{url}
\urldef{\mailsa}\path|{hallawa,ascheid}@ice.rwth-aachen.de|
\urldef{\mailsb}\path|{giovanniiacca}@gmail.com|
\urldef{\mailsc}\path|a.yaman@tue.nl|
\newcommand{\keywords}[1]{\par\addvspace\baselineskip
	\noindent\keywordname\enspace\ignorespaces#1}

\usepackage[acronym]{glossaries}
\newglossaryentry{kiea}{
	name = KIEA,
	description = Knowledge Integrated Evolutionary Algorithm
}

\newcommand{\cpop}[3][2]{|C^{#2}_{#3}|}
\newcommand{\clusterset}[4]{\mathcal{C}^{4}}
\newcommand{\ind}{\textbf{i}^{g}_{l}}
\newcommand{\cen}{\mathbf{c}^{g}_{l}}
\newcommand{\cenme}[3][2]{\mathbf{c}^{#2}_{#3}}
\newcommand{\indme}[3][2]{\textbf{i}^{#2}_{#3}}

\usepackage{amssymb}
\usepackage{algorithm}
\usepackage{algorithmicx}
\usepackage{algpseudocode}
\usepackage{bibnames}

\begin{document}
	
	\mainmatter

	\title{A Framework for Knowledge Integrated Evolutionary Algorithms}
	\titlerunning{Knowledge Integrated Evolutionary Algorithms}
    
     \author{Ahmed Hallawa\inst{1} \and Anil Yaman\inst{2} \inst{3} \and Giovanni Iacca\inst{2} \and Gerd Ascheid\inst{1} }

	
 	\institute{Chair for Integrated Signal Processing Systems\\
 			RWTH Aachen University, 52056, Aachen, Germany\\
 		\mailsa\\
        \and
         INCAS$^3$\\
 		P.O. Box 797, 9400AT, Assen, The Netherlands\\
 		\mailsb\\
         \and
         Department of Mathematics and Computer Science\\
 			Eindhoven University of Technology\\
 			P.O. Box 513, 5600MB, Eindhoven, The Netherlands\\
 		\mailsc\\
         }
	
	\toctitle{A Framework for Knowledge Integrated Evolutionary Algorithms}
	
    \maketitle
	\keywords{Evolutionary Algorithms, Knowledge Incorporation and EA Fingerprint}
	\begin{abstract}
        One of the main reasons for the success of Evolutionary Algorithms (EAs) is their general-purposeness, i.e., the fact that they can be applied straightforwardly to a broad range of optimization problems, without any specific prior knowledge. On the other hand, it has been shown that incorporating a priori knowledge, such as expert knowledge or empirical findings, can significantly improve the performance of an EA. However, integrating knowledge in EAs poses numerous challenges. It is often the case that the features of the search space are unknown, hence any knowledge associated with the search space properties can be hardly used. In addition, a priori knowledge is typically problem-specific and hard to generalize. In this paper, we propose a framework, called Knowledge Integrated Evolutionary Algorithm (KIEA), which facilitates the integration of existing knowledge into EAs. Notably, the KIEA framework is EA-agnostic (i.e., it works with any evolutionary algorithm), problem-independent (i.e., it is not dedicated to a specific type of problems), expandable (i.e., its knowledge base can grow over time). Furthermore, the framework integrates knowledge while the EA is running, thus optimizing the use of the needed computational power. In the preliminary experiments shown here, we observe that the KIEA framework produces in the worst case an 80\% improvement on the converge time, w.r.t. the corresponding ``knowledge-free'' EA counterpart.  
	\end{abstract}
        
	\section{Introduction}
    \label{intro}
	
    Evolutionary Algorithms (EAs) are considered nowadays a valuable search and optimization tool suitable for many real-world problems characterized by complex multidimensional search spaces. Among the many applications of EAs, some notable examples include the optimal design of electronic circuits \cite{koza2003s}, software \cite{arcuri2014co,squillero:2005:GPEM}, and even antennas for satellites orbiting outer space \cite{hornby2006automated,lohn2003evolutionary}.
	
	Despite the EAs' versatility, the theoretical limitations stated by the ``No free lunch" (NFL) \cite{wolpert1997no} pose a limit to their efficiency and applicability. As a possible mitigation for this problem, an EA can be made efficient and effective across a wide range of problems by endowing it with adaptive behavior with respect to the problem structure, thus with problem-specific mechanisms.
	
	Such adaptation typically involves the EA's operators and tunable parameters, which play a pivotal role in the performance of the algorithm. In fact, there are several methods that can be used to optimize the behavior of EA. In a recent survey published by {\v{C}}repin{\v{s}}ek et. al \cite{vcrepinvsek2013exploration}, a number of approaches that can be used for this purpose are presented. Traditionally, researchers have used trial-and-error approaches in the attempt to find the best settings of the EA operators that can solve optimization problems most efficiently \cite{back1994selective,gates1995simple}. However, these approaches are typically computationally expensive, because they require numerous iterations (and in many cases it is not feasible to try all possible parametric combinations), and some are problem-specific, thus they can not be used on problems other than the one on which the tuning was performed. Furthermore, proposed frameworks which offer an adaptive behavior such as in some modulated versions of Differential Evolution (DE) algorithms, as in jDE \cite{yang2013improved} and JADE \cite{caorsi2005optimization}, doesn't offer a comprehensive strategy for all tunable parameters. Other approaches uses hyper-heuristics, i.e. they find the optimal EA settings by using an optimization algorithm \cite{vcrepinvsek2013exploration,Eiben1999}. 
    
	One of the approaches that have not yet been explored, however, is to use experiences from previous problems with similar population behavior in the EA run \cite{vcrepinvsek2013exploration}. Here, we propose a framework that is a first attempt in this direction, where we also combine the approaches of ``following general guidelines" accumulated in the literature, and ``identifying the features of the landscape by a classifier, in order to propose good control parameters" (see \cite{vcrepinvsek2013exploration}). 
    
	The proposed framework is dubbed as \emph{Knowledge Integrated Evolutionary Algorithm} (KIEA). Its main component is a \emph{knowledge base} that maintains the knowledge of how various functions (i.e., optimization problems) can be efficiently solved. These functions are named as \emph{pilot functions}, and the associated knowledge to optimally tune the EA for solving those functions is named a \emph{strategy}. The framework collects characteristics of the EA population behavior across generations. These characteristics are named \textit{EA fingerprints}, and they are used to classify any unknown function under investigation w.r.t. each of the pilot functions (under the implicit hypothesis that such fingerprints can be used to assess the similarity between different functions). The strategy associated to the classified pilot function is then reused on the unknown function at hand, in the attempt of solving it in the most efficient possible way. 
	
	The proposed approach is novel in several ways. Firstly, it allows the incorporation of various types of knowledge into the EA, allowing the algorithm to adjust its behavior based on the knowledge in the knowledge base. Initially, experts can bootstrap the system with their knowledge on the pilot functions. Secondly, the experience gained by the EA by solving problems generates valuable empirical knowledge that can also be added to the knowledge base for further use. This can be done by extending the pilot functions or changing the strategies associated with them. As a result, the knowledge base grows by accumulating the experience gained by solving multiple problems. The accumulated knowledge is then ``plugged'' into the problems that are similar to the ones encountered before. Moreover, the KIEA approach is generalizable as it conducts the problem classification only based on the behavior of the population in the EA run, independently from what this population is representing, i.e. the solution encoding and the genotype/phenotype mapping.
	
    To assess the performance of KIEA, we perform preliminary experiments on a small set of benchmark optimization problems. More specifically, we measure the fingerprint-based classification accuracy on different pilot functions by using different fingerprint properties. In addition to that, we compare the difference in terms of convergence time obtained in the experiments with and without KIEA.
    
	The rest of the paper is organized as follows: Section \ref{background} summarizes the previous works on knowledge integration in EAs, Section \ref{methods} presents the mathematical foundation of the EA fingerprint and the classification process. Section \ref{results} shows the numerical results. Finally, Section \ref{conclusions} concludes the paper. 

	\section{Background}
    \label{background}
	
    Different knowledge incorporation methodologies available in the literature aim to optimize EAs in order to enhance their performance. One of the key elements where knowledge plays a role is the balance between exploration and exploitation, a crucial aspect for an efficient search \cite{vcrepinvsek2013exploration,eiben1998evolutionary}. If there is, for example, more influence of exploration then the search becomes more like a random search; on the other hand, if exploitation is stronger than exploration then the search space could not be explored, and the behavior of the search becomes similar to the behavior of hill climbing \cite{vcrepinvsek2013exploration}.
	
	In EAs, the balance between exploration and exploitation is typically adjusted by the evolutionary operators and their parameters. However, different evolutionary operators and different parameter values influence the process differently; and their combinations may have different, hard-to-analyze effects. One general interpretation considers the mutation and crossover as exploration operators\footnote{It should be noted, however, that some literature considers the crossover operator as an exploitation mechanism. Generally, mutation and crossover have an effect on both exploration and exploitation, although this effect varies depending on the implementation and the fitness landscape at hand.}, since they make (pseudo-)random changes in the genotype of individuals and cause random jumps in the search space; on the other hand, the selection operator is is usually seen as an exploitation operator, because it focuses on specific places by selecting the individuals to reproduce. Moreover, the population size plays an important role in the EA behavior. Generally, re-sizing population can be used to direct evolution process towards exploration or exploitation \cite{harik1999parameter,harik1999compact,iacca2011super}, this method is widely used in CMA-ES, DE, and PSO. In that regard, it is also important to highlight that the increase of the population size does not necessarily improve exploration. Conversely, it has shown that there is strong link between population size and structural bias of the algorithm \cite{kononova2015structural}. Consequently, this component has to be taken into consideration when designing self-adapting EA.
	
	The performance of a search process is also closely dependent on the features of the search space and the fitness landscape. There have been many works in the literature that aim to classify the landscapes based on their geometrical and topological features. Most of these works are linked to specific features such as modality, symmetry, etc., such as in \cite{beyer2002evolution}, \cite{casas2015genetic}, \cite{miller1996genetic}, and \cite{sareni1998fitness}. There have also been several studies that aim to suggest optimal algorithms, or optimal algorithm parameters, based on the features of the landscape of the search space. For example, Asmus et al. \cite{asmus2014towards} proposed a system for recommending suitable algorithms for a given black-box optimization problem. Muñoz et al. \cite{munoz2012meta} introduced a model that links the landscape analysis measures and the algorithm parameters (used CMA-ES) to predict the performance of the algorithm parameters. Picek and Jakobovic \cite{picek2014fitness} performed a thorough study focused on the correlation between fitness landscapes and crossover operators. 
    
	Clearly, it would be extremely beneficial to leverage this wide range of knowledge from the literature for tuning the evolutionary operators parameters based on the landscape features. However, applying this knowledge often requires that such features are captured first, in order to choose the proper strategies. Unfortunately though, this is not always possible since either the search space is completely unknown, or hard to characterize.
	
	Moreover, many of the existing works offer knowledge that is problem-specific and therefore hard to generalize and use with other problems. Another difficulty arises from the fact that this knowledge is often scattered over different levels of granularity, from too general to extremely detailed, which makes it hard to have comprehensive strategies. 

    The objective of the present work is to propose a way of integrating existing \emph{algorithmic knowledge} into a single, comprehensive evolutionary framework, and test the effect of such knowledge on the optimization performance. Here, with ``algorithmic knowledge'' we generally refer to the knowledge encompassing the categorizations of problems based on their landscape features, the types of strategies related to the adaptation of evolutionary operators and parameters, and the link between problem types and strategies, i.e. which strategies work best on a specific type. In the next section a detailed description of the framework is presented.
	
	\section{Methods}
	\label{methods}
    
\begin{figure}[!ht]
\centering
\includegraphics[ scale = 0.35 ]{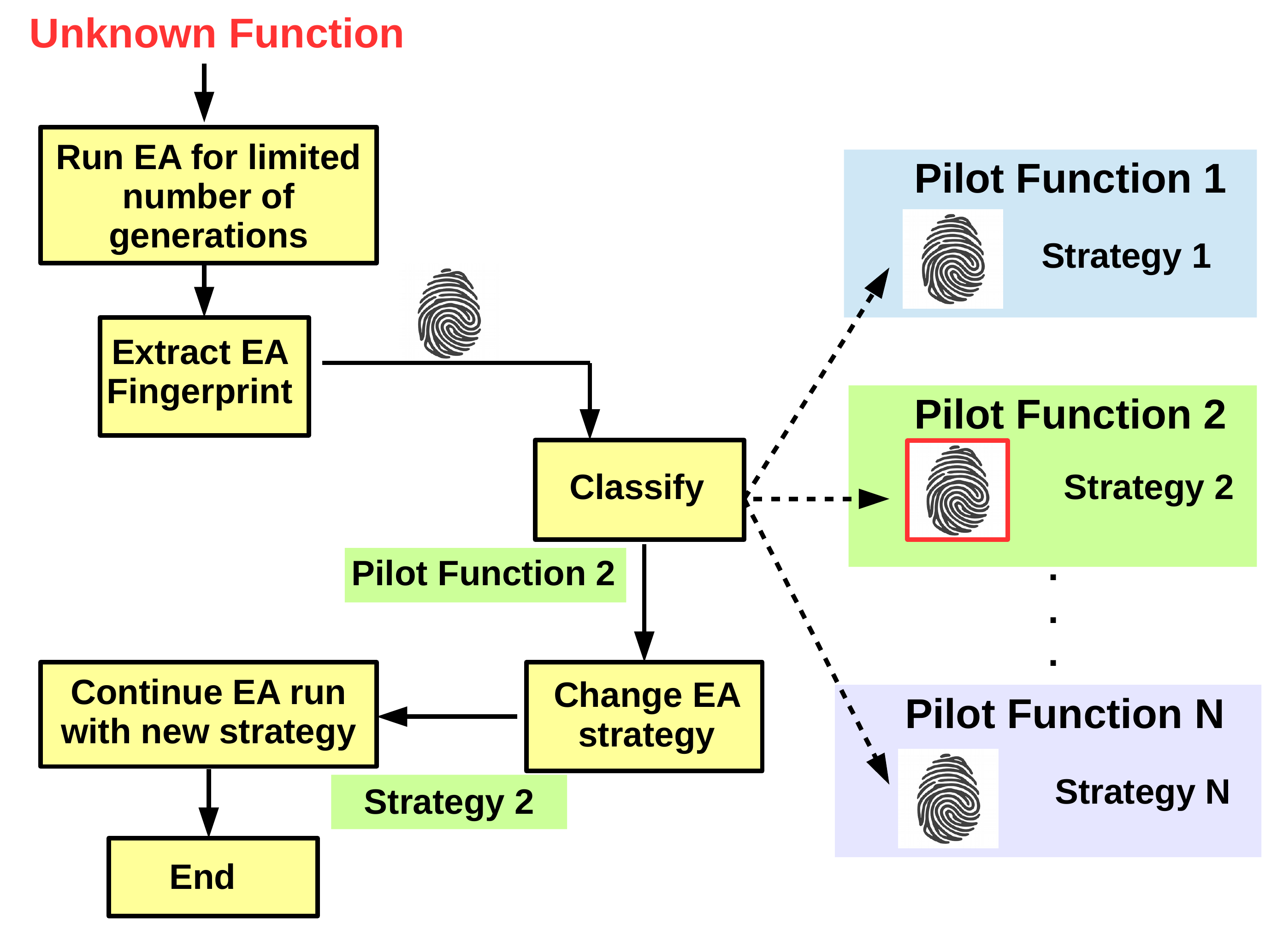}
\caption{Conceptual scheme of KIEA}
\label{Layout}
\end{figure}

    The KIEA methodology is straightforward and its implementation is relatively simple (see Algorithm \ref{KIEA procedure}). For completeness, we also report a conceptual scheme of the proposed KIEA in Figure \ref{Layout}.

    The whole process can be considered as a single run of an EA divided into two stages: in the \textbf{first stage}, for a predefined number of generations, $G_C$, the unknown function under investigation undergoes an EA run with an arbitrary parameter setting (basically, population size, mutation and crossover probability). These settings are set as the initial EA strategy $S_0$. In each generation evolved in this first stage, a set of properties describing the population behavior is calculated to be used later for classification. These population behavior properties are termed \textit{EA fingerprint} (see Section \ref{fingerprint} for details). The first stage stops when the allotted number of generations $G_C$ is reached.
	
	In the \textbf{second stage}, a classification based on the EA fingerprint is performed as follows. The unknown function's fingerprint produced from the first stage is compared with the fingerprint obtained by the the same initial strategy $S_0$ on a set of \emph{pilot functions}, i.e. benchmark functions that are chosen a priori as representative of different categories of problems. Based on fingerprint similarities, the unknown function is then classified as the most similar pilot function. Detailed insights on the classification process are provided in Section \ref{classification}. 
    
    Associated to each pilot function, the system maintains an EA \emph{strategy}. Here, we refer to ``strategy'' as a set of changes (adaptation rules) in the EA parameters (e.g. a population size reduction/shrinking, mutation probability update rules, etc.) and when these changes should be implemented within the evolutionary run in order to enhance the optimization performance. These strategies are organized in a \emph{knowledge base}, such that for each pilot function, its associated strategy is the one that according to our empirical experiments (see Section \ref{strategy}) showed the best performance.
    
	Therefore, after the classification phase, the settings of the algorithm (i.e. mutation rate, population size, etc.) are set according to the EA strategy that is associated with the most similar pilot function. For the remaining of the available generations, the EA runs with the chosen strategy. The hypothesis is that similar strategies can enhance the performance of an EA on functions with similar EA fingerprints; in other words, we expect the performance of the optimization process to improve on the unknown function after adopting the strategy associated with the most similar pilot function.
	
	It is important to notice that the ability to classify problems based on the EA fingerprint makes it possible to transfer the knowledge from the pilot functions to any unknown function under investigation, without assuming any previous understanding of its features or properties. Moreover, this mechanism allows generalization of knowledge to a wide range of optimization problems without the need of associating such knowledge to specific fitness landscape properties such as modality, symmetry, etc. Therefore, it avoids the complexity due to landscape analysis, and makes it possible for the system to work with functions with complex, hard-to-analyze landscapes, since all that is required is to capture the population behavior in the EA run. 
	
	\begin{algorithm}[!ht]
		\begin{algorithmic}[1]
			\Procedure{KIEA}{}
			\State initialize total no. of generations $G$
			\State initialize no. of generations for classification $G_C$
			\State set initial EA strategy $S_{0}$
			\State initialize population $P$
			\While{$g < G_C$} 
			\State$F\gets$ evaluate $(P)$
			\State$P\gets$ select $(P,F)$
			\State $P\gets$ reproduce $(P,F,S_0)$
			\State $f \gets$ getFingerprint$(P,F)$ \Comment{Store fingerprint}
			\EndWhile 
			\State $Pilot_{i}\gets$ classify$(f)$ \Comment{Classification}
			\State $S_{i}\gets$ getStrategy($Pilot_{i}$) \Comment{Retrieve strategy}
			\While{$g < R_T$}
			\State$F\gets$ evaluate $(P)$ 
			\State$P\gets$ select $(P,F)$
			\State $P\gets$ reproduce $(P,F,S_{i})$
			\EndWhile
			\EndProcedure
		\end{algorithmic}
		\caption{High-level description of the KIEA framework}
		\label{KIEA procedure}
	\end{algorithm}
	
    In the following sections, we cover the mathematical details of the EA fingerprints and the classification procedure. In Table \ref{symbols}, we summarize the main symbols used in the text, with the related explanation.
  
\begin{table}
\caption{Symbols used in the paper}\label{symbols}
\begin{center}  
\begin{tabular}{p{1.9cm}l} 
\textbf{Symbol} & \textbf{Explanation}\\
\midrule
$N$ & Total number of individuals in population\\
$G$ & Total number of generations \\
$G_C$ & Number of generations allocated for classification\\
$\ind$ $\in$ $\mathbb{R}^n$ & Individual belongs to cluster $l$ at generation $g$\\
$\cen$ $\in$ $\mathbb{R}^n$ & Cluster $l$ center individual at generation $g$\\ 
$c_\text{min}$ & Minimum number of individuals to form a cluster\\
$r_\text{max}$ & Maximum radius of the sphere $S^{n}$ that a cluster can occupy\\
$ C^g_l$ & Set of individuals in cluster $l$ at generation $g$\\
$ \cpop{g}{l}$ & Number of individuals in cluster $l$ at generation $g$ \\
$ C^g_c$ & Set of cluster centers at generation $g$\\
$\mathcal{P}_l$ & A tuple with the number of individuals in cluster $l$ \\ 
$ d_{ij}^g$ & Euclidean distance between $\cenme{g}{i}$ and $\cenme{g}{j}$ at generation $g$\\
$T_l$ & Population trend tuple of cluster $l$\\
$ G^g_{d_{ij,\epsilon}}$ & Set of points that have equal Euclidean distance $d_{ij} \pm \epsilon$ at generation $g$\\
$S_i$ & EA Strategy $i$\\
\midrule
\end{tabular}
\end{center}
\end{table}
    
	\subsection{EA fingerprint}
	\label{fingerprint}

In the previous section we defined the EA fingerprint as a set of properties that characterize the population behavior. While in principle this could be done at individual level, in practice following all individuals in the population would be extremely computationally expensive, especially in high dimensions. For example, a simple task as finding the pair of points in a set with smallest distance between them (known as \textit{closest pair of points problem}) has time complexity of $O(n^2 D/log^2 D)$ for $D$ dimensions and $n$ points using a divide and conquer approach \cite{min2009closest}. Therefore, we define here EA fingerprints that are based on \emph{clustering} the population and following the resulting clusters properties. 
	
	A cluster emerges when a predefined minimum number of individuals $c_{min}$ from the population are grouped in a predefined maximum space $r_{max}$ in the search space. This minimum number of individuals constituting a cluster and the corresponding maximum space used to designate it are set as a percentage of the population size $N$, e.g $c_{min}$ = 5\% of $N$. 
	
	An EA fingerprint is grouped into two main groups: Clusters Emergence Characteristics ($\operatorname{CEC}$) and Clusters Constellation Characteristics ($\operatorname{CCC}$). $\operatorname{CEC}$ is designed to capture 5 features: number of clusters, number of individuals in each cluster, population trend in each cluster, fitness value of the fittest individual of each cluster, and its position. On the other hand, $\operatorname{CEC}$ captures the geometric properties of clusters, which include equidistant cluster topology and the corresponding distances between equidistant clusters in the search space. 
	
	In $\operatorname{CCC}$, the first step is recognizing emerging clusters. The procedures for that are described as follows:
	\begin{enumerate}
		\item All individuals $\ind$ $\in$ $\mathbb{R}^n$ $\forall \ l = 1\ldots N$ at generation $g$ are sorted in descending order with respect to fitness.
		\item The highest value is designated as a potential cluster center $\cen$ forming the first point in the potential cluster set $C^g_l$.
		\item Going through all population, each individual $\ind$ is assigned to $ C^g_l$ with center point $\cenme{g}{i}$ if and only if:
		\begin{equation}
			\Vert \indme{g}{i} - \cenme{g}{j} \Vert < r_{max}, \quad \forall j \neq i \ \wedge \ j = 1\ldots N
		\end{equation}
		where $r_{max}$ is the maximum radius of the sphere $S^{n}$ that a cluster can occupy, $S^{n}$=$\{ x \in \mathbb{R}^{n+1} : \Vert x \Vert = r_{max}$\}. $r_{max}$ is chosen adequately, e.g. 5\% of the smallest search domain across all search variables.
		\item A cluster $ C^g_l$ is designated with the center $\cenme{g}{i}$ if and only if the number of individuals assigned to it, $\cpop{g}{l}$, is bigger than or equal to the minimum cluster size $c_{\text{min}}$:
		\begin{equation}
			\cpop{g}{l} \geq c_{\text{min}}
		\end{equation}
		where $c_{\text{min}}$ is chosen as a percentage of the total population size $N$, e.g. 5\%.
		
		\item All individuals that were previous assigned to a cluster or picked as a potential cluster center are then discarded, steps (3) to (5) are repeated again until all $N$ individuals in the population are considered.
	\end{enumerate}
    
	These procedures are executed for each generation, until $G_C$ is exhausted. For each generation $g$, all cluster centers $\cen$, $l=1,2,\dots$, are kept in a set $C^g_c$. Furthermore, due to sorting population in the first step in the procedures, $\cen$ are also the fittest points in cluster $l$. This will be used later for comparing clusters with similar highest fitness points. In addition, for each generation the corresponding fitness value of each cluster center $\cen$ are stored in a fitness set $F^{g}_c$.
	A tuple $\mathcal{P}_l$ with the number of individuals in cluster $l$ through out all generations until a given generation $m$ is defined as follows:
	\begin{equation}
		\mathcal{P}^m_l = <  \cpop{1}{l}, \cpop{2}{l}\ldots, \cpop{\text{m}-1}{l}, \cpop{\text{m}}{l} > \;
	\end{equation}
	where $m$ can take any value from 1 to total number of generations $G$. In order to capture the changes in each cluster throughout different generations until a given generation $m$, for each $\mathcal{P}^m_l$, there exists a population trend tuple $T^m_{l}$ defined as:
	\begin{equation}
		\label{tuple}	
		T^m_{l}=
		\begin{cases}
			1,&  \text{for} \;  \cpop{k}{l} -  \cpop{k+1}{l} < 0\\
			0,&  \text{for} \; \cpop{k}{l} -  \cpop{k+1}{l} > 0
		\end{cases}	
		\qquad \forall k = 1\ldots m-1
	\end{equation}
	Now, all the aforementioned $\operatorname{CEC}$ properties can be defined: the number of clusters for each generation $g$ in $\cpop{g}{l}$, the number of individuals in each cluster $l$ in $\mathcal{P}_l$, the population trend across all generations in each cluster in $T_{l}$ , the fitness value of the fittest individual (which is also the center) of each cluster in $F^{g}_c$, and its position in $C^g_c$ for each generation $g$.
	
	The second fingerprint component's, $\operatorname{CCC}$, is meant to capture the geometric properties of clusters. Firstly, we define the Euclidean distance between all cluster centers in each generation $g$ as:
	\begin{equation}
		\operatorname{d}^{g}_{ij} =  \Vert \cenme{g}{i} - \cenme{g}{j}  \Vert 
	\end{equation}
Then, we group equidistant cluster center points as follows:
	\begin{equation}
		\operatorname{G}^g_{d_{kl,\epsilon}} = \{ (\cenme{g}{i},\cenme{g}{j}) \;\in C_{c}^{g^2}   \;  \forall i\neq j, \; d_{kl}-\epsilon < \Vert \cenme{g}{i} - \cenme{g}{j}  \Vert < d_{kl}+\epsilon \}
	\end{equation}
	where $\epsilon$ is a margin of tolerance adequately chosen as a percentage of the domains of each search variable.
	
	Both $\operatorname{d}^{g}_{ij}$ and $\operatorname{G}_{d_{kl,\epsilon}}$ constitute the $\operatorname{CCC}$ properties. Now that we have defined all the elements of the EA fingerprint, we can show how we use them for classification.
	
	\subsection{Classification}
	\label{classification}
    
	The objective of the classification process is to find the closest pilot function to the unknown function under investigation. This is conducted by comparing the EA fingerprints with the fingerprint of each pilot function. In the following description, $\psi^g_k$ indicates the comparison of feature $k$ at generation $g$. 
	
	The first comparison $\psi^g_1$ is the difference in cluster numbers at each generation. This difference is multiplied by the ratio between the smallest cluster number over the biggest, as follows:
	\begin{equation}
		\psi_1  = \Vert {\cpop{g}{l}}_{u} - {\cpop{g}{l}}_{p} \Vert \times \frac{\text{min}({\cpop{g}{l}}_{u} ,  {\cpop{g}{l}}_{p})}{\text{max}({\cpop{g}{l}}_{u} , {\cpop{g}{l}}_{p})}
	\end{equation}
	where ${\cpop{g}{l}}_{u}$ and ${\cpop{g}{l}}_{p}$ are the numbers of clusters on the unknown function and the pilot function, respectively.
    
	The second comparison $\psi^g_2$ is the difference in the position between each center point $\cen$ in the the unknown function and the closest cluster center point position in the pilot functions. Only points whose distance from $\cen$ is at most $r_{max}$ are considered, to ensure that the closest point found in the pilot function cannot be chosen more than once, since there cannot exist two center cluster points within the same cluster sphere $S^{n}$ which has maximum radius $r_{max}$. Consequently, $\psi^g_2$ is defined as the sum of all the minimum distances between $\cen$ of unknown function and pilot function, as follows: 
	\begin{equation}
		\label{et1}
		\psi^g_2 = \sum_{\cenme{g}{i} \in {^u\text{C}_l^g}} \; \underset{\cenme{g}{j} \in {^p\text{C}_l^g}}{\mathrm{min}} \Vert \cenme{g}{i} - \cenme{g}{j}  \Vert \; \text{with} \;d_{ij} < r_{max} 
	\end{equation}
	where $^u\text{C}_l^g$ and $^p\text{C}_l^g$ are sets with center points $\cen$ at generation $g$ for the unknown function and pilot function, respectively. The inequality $d_{ij} < r_{max}$ is the minimum distance condition explained earlier.
	
	The third comparison is defined to capture the difference in population trends between the unknown function and the pilot function for clusters with similar fittest point values, i.e. clusters which have close center fitness values across all generations. Center cluster points with similar fitness values are identified as follows:
	
	\begin{equation}
		\label{et2}
		^uT^g_{f_0+k\epsilon} = \{ f_l \in F^g_u|f_l - \epsilon \leq f_0 + k\epsilon \leq  f_l + \epsilon \}  \; \forall k \in \mathbb{N}
	\end{equation}
	where $f_0$ is the least fit cluster center $\cen$ in the unknown function at generation $g$, and $F^g_u$ is the fitness set of the unknown function at generation $g$. $^uT^g_{f_0+k\epsilon}$ is a set of all points within range $f_0+k\epsilon$, $\forall k \in \mathbb{N}$. The analogous value for the pilot functions, $^pT^g_{f_0+k\epsilon}$ is calculated similarly:
	\begin{equation}
		^pT^g_{f_0+k\epsilon} = \{ f_l \in F^g_p|f_l - \epsilon \leq f_0 + k\epsilon \leq  f_l + \epsilon \} \; \forall k \in \mathbb{N}
	\end{equation}
	where $f_0$ is least fit cluster center $\cen$ in the unknown function at generation $g$ (same as in equation \ref{et2}), and $F^g_p$ is the fitness set of the pilot function at generation $g$. Then $\psi^g_3$ is defined as:
	\begin{equation}
		\psi^g_3  =  \sum_{k\in \mathbb{N}} \frac{\min(|^uT^g_{f_0+k\epsilon}|, |^pT^g_{f_0+k\epsilon}|)}{\max(|^uT^g_{f_0+k\epsilon}|, |^pT^g_{f_0+k\epsilon}|)}\quad \times \frac{\min({\cpop{g}{l}}_{u} , {\cpop{g}{l}}_{p})} {\max({\cpop{g}{l}}_{u} , {\cpop{g}{l}}_{p})} 
	\end{equation}
	where $\psi^g_3$ is the summation of the ratio of number of points in range $f_o + k\epsilon$ in the unknown function and pilot function $\forall k$, normalized w.r.t. the ratio of cluster sizes, exactly like in $\psi^g_1$. The reason why the $\min$ and $\max$ operators are used instead of simply having $|^uT^g_{f_0+k\epsilon}|$ in the numerator and $|^pT^g_{f_0+k\epsilon}|$ in the denominator, is to ensure that the value of the fraction is always less than one and therefore $\psi^g_3 $ is comparable with other cases, regardless the fact that $|^uT^g_{f_0+k\epsilon}|$ is greater than $|^pT^g_{f_0+k\epsilon}|$ or vice versa.
    
The fourth comparison is used to capture the difference between the equidistant cluster center points, as follows:    
	\begin{equation}
		\psi^g_4  = \frac{\min(|\operatorname{G}^g_{d_{ij,\epsilon}}|_{u} , |\operatorname{G}^g_{d_{kl,\epsilon}}|_{p})}{\max(|\operatorname{G}^g_{d_{ij,\epsilon}}|_{u} , |\operatorname{G}^g_{d_{kl,\epsilon}}|_{p})} \; \text{with} \; d_{kl} - \epsilon \leq d_{ij} \leq d_{kl} + \epsilon
	\end{equation}
	where $\operatorname{G}^g_{d_{ij,\epsilon}}$ is the set of equidistant cluster centers within the margin of tolerance $\epsilon$ at generation $g$.
    
	The fifth comparison captures the trend of the population within  clusters. Its definition reflects similarities in the change of population within clusters with similar fitness values.   	
	$f_l^g$ is the highest fitness point in each cluster $l$ at generation $g$, and there might exist a $f_l^g$ in the pilot function at the same generation $g$ with fitness value that is within $\epsilon$ range from it, a set that includes all these pairs of points is defined as:  
	\begin{equation}
		M^g = \{ (i,j) \in \mathbb{N}^2| \; \forall f_i \in F^g_u, f_j \in F^g_p, \,\; \Vert f^g_i - f^g_j \Vert \leq  \epsilon\} 
	\end{equation}
	where $F^g_u$ and $F^g_p$ are the fitness sets of the unknown function and pilot function respectively. $M^g$ includes all cluster id pairs unknown function and pilot function that have highest fitness values difference less than $\epsilon$. From equation \ref{tuple}, each item at position $k$ in the tuple reflects the change in population between generation $k$ and $k+1$, its value be either zero for no change in population or one for increase in population or negative one for a decrease in population. For each pair identified in$ M^g$, a comparison in the population number history up to generation $g$ is defined as follows:
	\begin{equation}
		\label{et1}
		\psi^g_5 =  \sum_{(i,j)\in M} T^g_i \circledcirc T^g_j
	\end{equation}
	where $T^g_i$ and $T^g_j$ are the trend tuples of the unknown function and pilot function respectively and $\circledcirc$ is defined here as an XNOR logic operator which produces 1 if operands are equal and 0 otherwise.
    
    The sixth comparison captures the difference in the number of clusters per generation, as follows:
	\begin{equation}
		\psi^g_6  =  \sum_{g=1}^{G}\frac{\text{min}({\cpop{g}{l}}_{u} ,  {\cpop{g}{l}}_{p})}{\text{max}({\cpop{g}{l}}_{u} , {\cpop{g}{l}}_{p})}
	\end{equation}
    
	Finally, the classification process is concluded based on $\psi_1$ to $\psi_6$, using a weighted sum as follows:
    \begin{equation}
		\psi_{Total}  =  \sum_{i=1}^{6} w_i\psi_i
	\end{equation}
    where $w_i$ is the weight of property $i$, which is set as follows:
    \begin{equation}
		\label{tuple}	
		w_i=
		\begin{cases}
			 1,&  \text{for} \;  i = 3,\\
			-1,&   \; otherwise
		\end{cases}	
		\qquad \forall i = 1\ldots 6
	\end{equation}
as with exception to $\psi_3$, the lesser the value the better the match with the pilot function. Finally, the total value is then compared with each pilot function, and the highest value leads to the winning pilot function, which then leads to the unknown function adapting its strategy.
	
	\section{Results}
	\label{results}
    As a proof-of-concept, we now present the numerical results obtained by KIEA on a small set of benchmark functions. First, we explain the algorithmic setup and the strategy initialization in Section \ref{strategy}. Then in Section \ref{fingerprint results} we illustrate the effect of different fingerprint properties on classification. Finally, Section \ref{KIEAPerformance} shows the KIEA performance in terms of convergence time.
     
    \subsection{System setup and strategy initialization}
	\label{strategy}
	
\begin{figure}[!ht]
	\centering
		\includegraphics[ width=\linewidth]{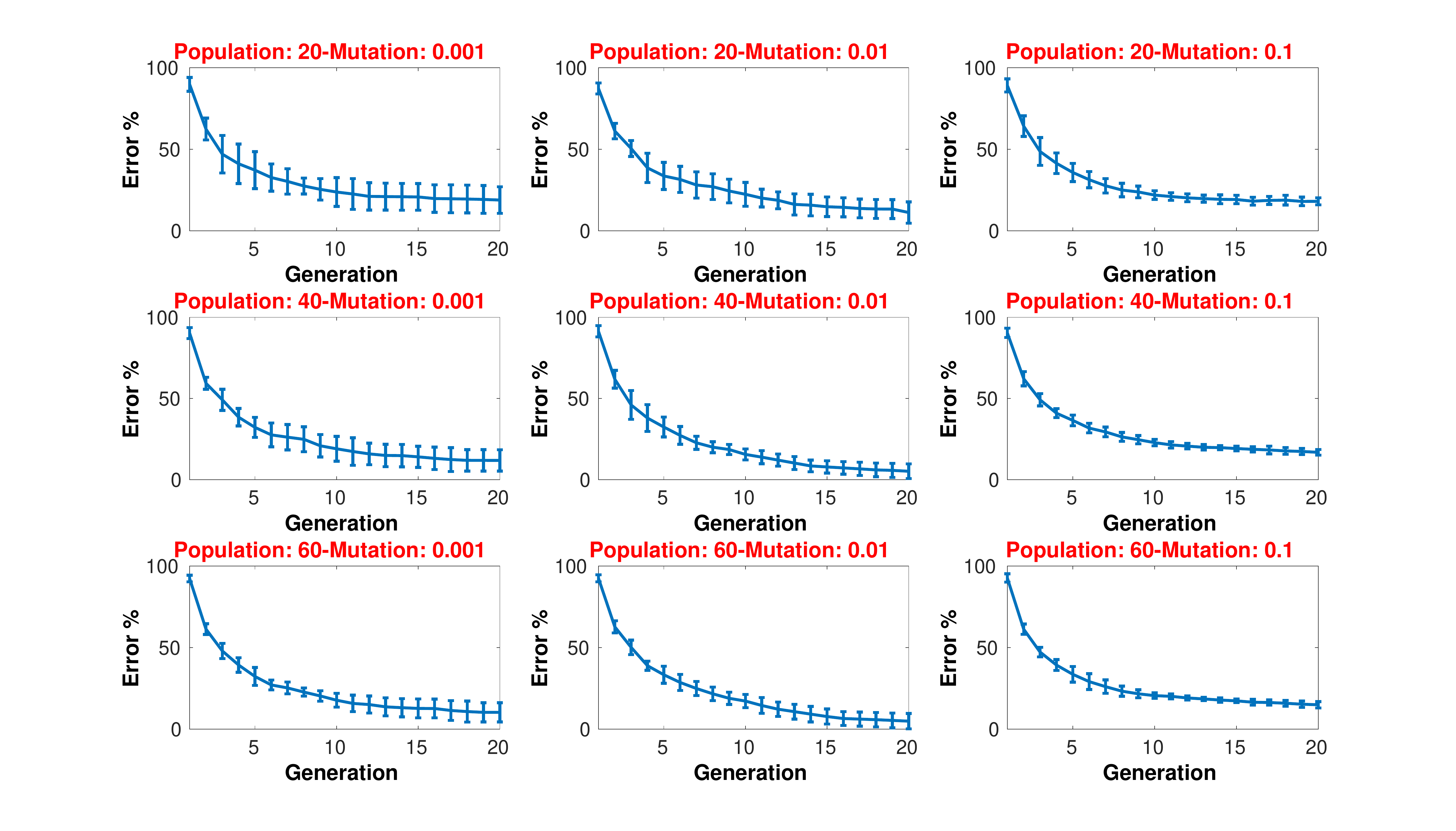}
		\caption{Strategy analysis on the Ackley function}
	\label{ackley}
	\end{figure}

\begin{figure}[!ht]
	\centering
		\includegraphics[ width=\linewidth]{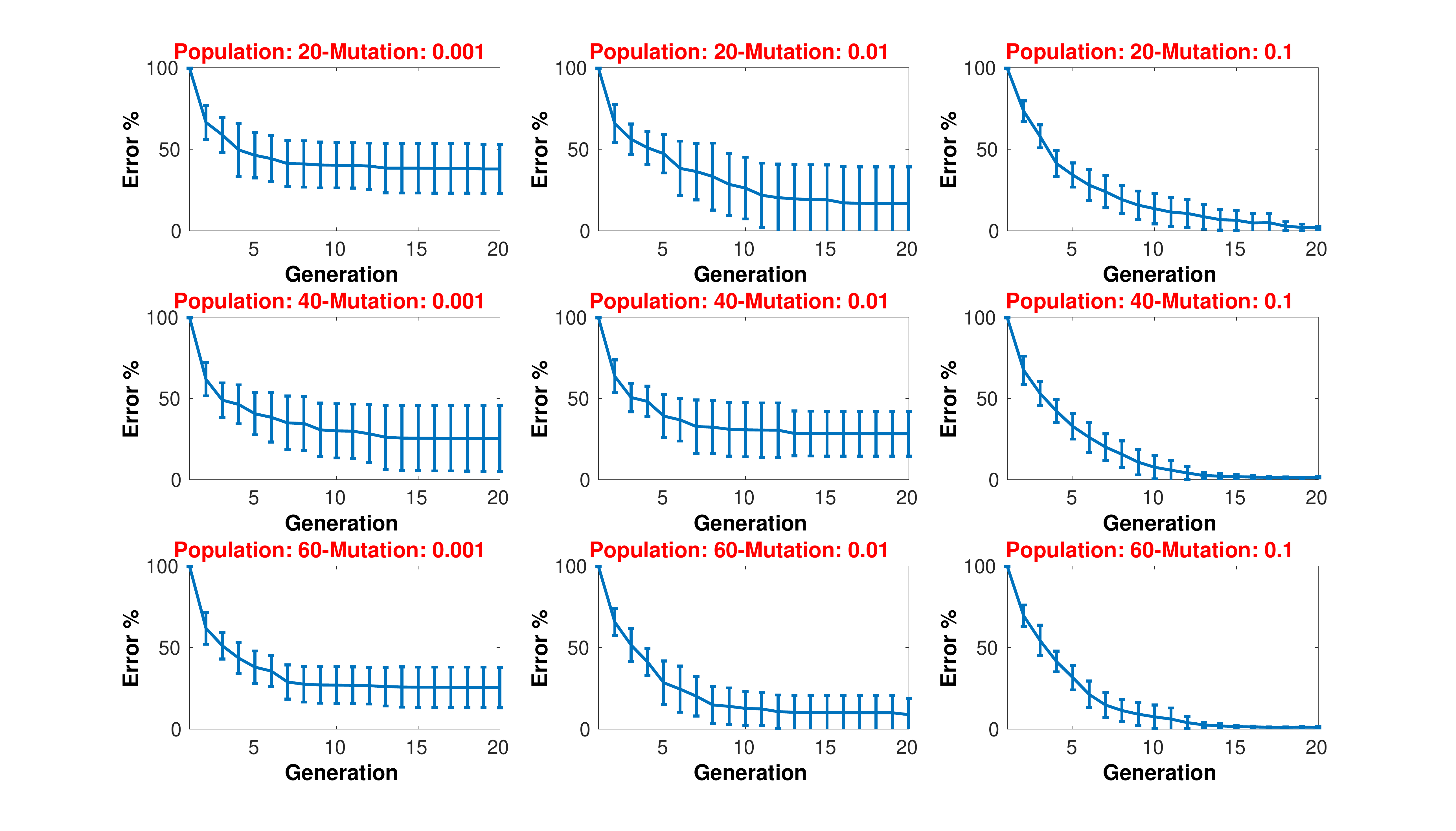}
		\caption{Strategy analysis on the Gaussian function}
	\label{gaussian}
	\end{figure}

	Although the KIEA framework is algorithmically agnostic and can be used with any EA, for testing purposes we use a classic Genetic Algorithm (GA). In the prototype we implemented, we focused on strategies expressed in terms of population size and mutation rates. Figures \ref{ackley} and \ref{gaussian} show a preliminary performance analysis for two simple benchmark functions, namely the Ackley and the Gaussian function. It can be seen that for the Ackley function, population size $60$ and mutation rate $0.01$ is the most suitable strategy, while for the Gaussian function population size $40$ and mutation rate $0.1$ offer a more suitable strategy. 
    
    \subsection{Effect of fingerprints}
    \label{fingerprint results}
    
\begin{figure}
   \centering
   \includegraphics[width=\linewidth]{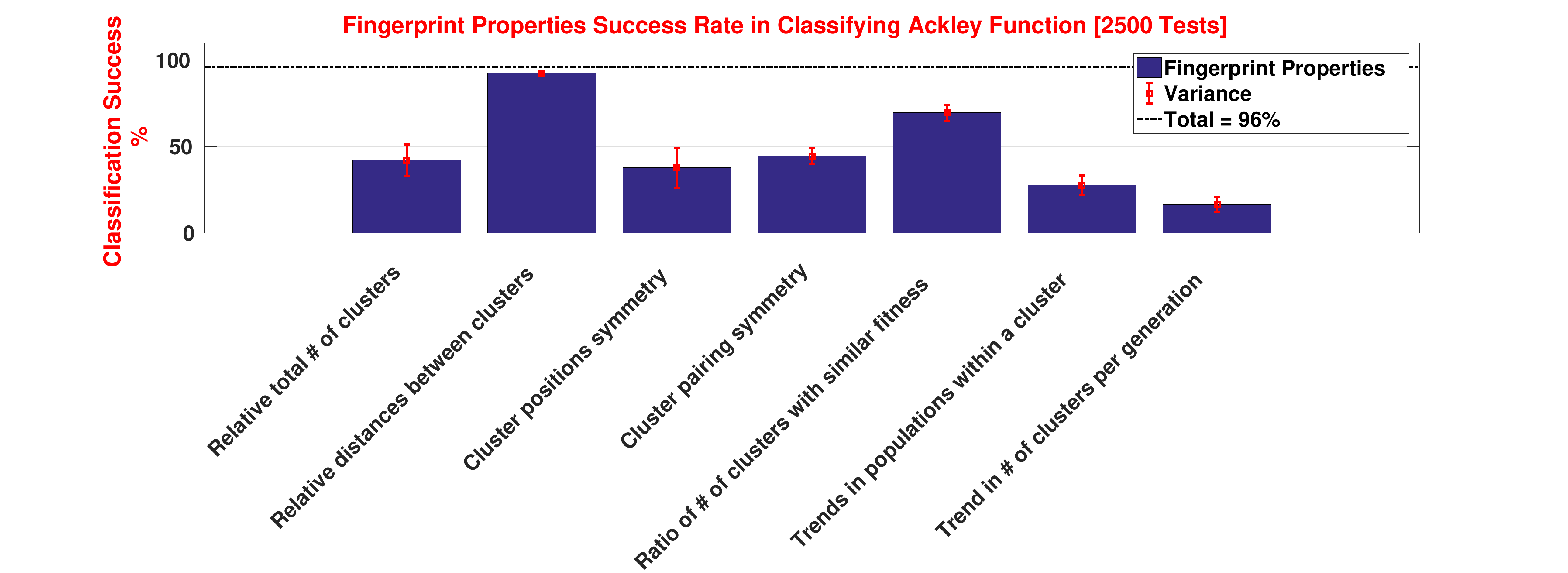}
   \caption{Classification of the Ackley function}
   \label{ackleyClassification}
   \vspace{-0.8cm}
   \end{figure}

\begin{figure}
   \centering
   \includegraphics[width=\linewidth]{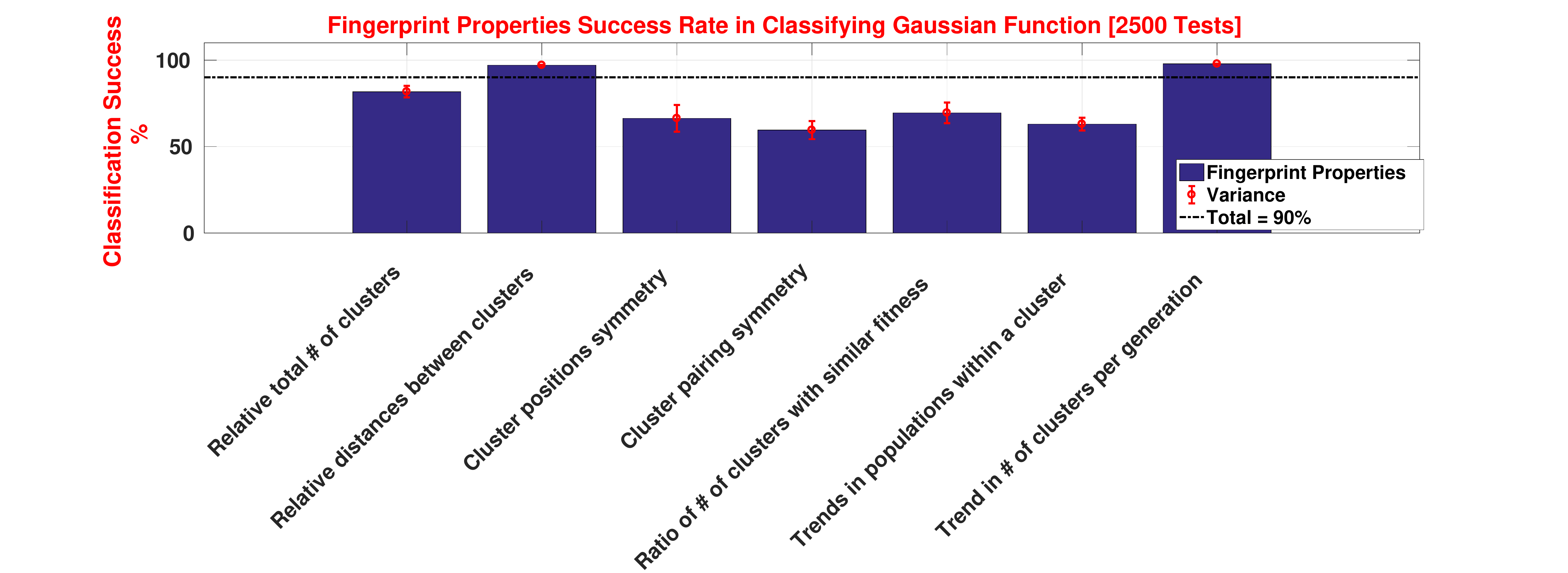}
   \caption{Classification of the Gaussian function}
   \label{gaussianClassification}
   \vspace{-0.8cm}
   \end{figure}

    To test the ability to classify a function based on its fingerprint, we conducted a four tests in total. In the first two, the objective was to classify the two pilot functions (Ackley and Gaussian) as if they were unknown. Moreover, we considered two additional unknown functions (Rastrigin and Rosenbrock), different from the pilot functions. Figures \ref{ackleyClassification}-\ref{gaussianClassification} show the classification of the Ackley, Gaussian, Rastrigin, and Rosenbrock functions, respectively. Each classification test was done by first extracting 50 different fingerprints for each pilot function, and then comparing each of the 50 fingerprints from the unknown function, thus with a total of 2500 comparisons (50 pilot functions fingerprints $\times$ 50 unknown function fingerprints). It is important to highlight that each property in the fingerprint contributes to the classification process, i.e they are all needed and their contribution varies depending on the function at hand, while keeping the overall successful classification rate 90\% to 98\%. Moreover, their variance is relatively small (3\% on average), which shows the good reliability of the proposed classification process. 
   
   \begin{figure}
   \centering
   \includegraphics[width=\linewidth]{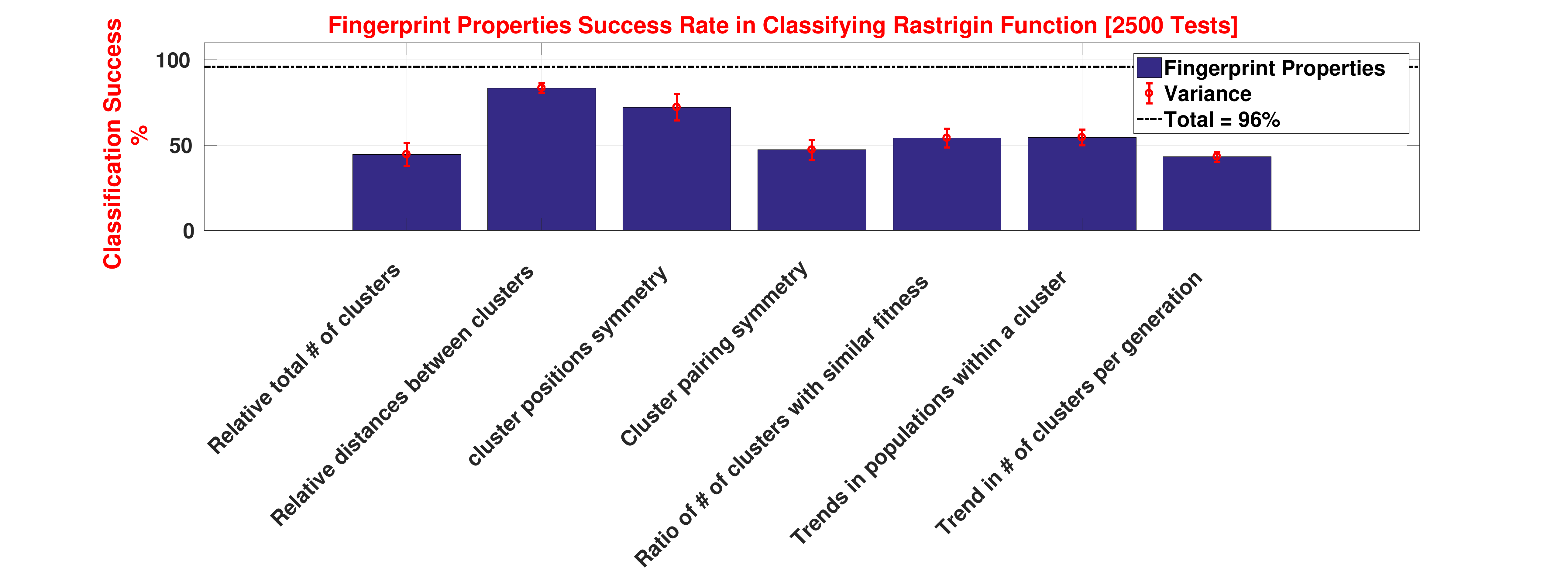}
   \caption{Classification of the Rastrigin function}
   \label{rastriginClassification}
   \vspace{-0.5cm}
   \end{figure}
   \begin{figure}
   \centering
   \includegraphics[width=\linewidth]{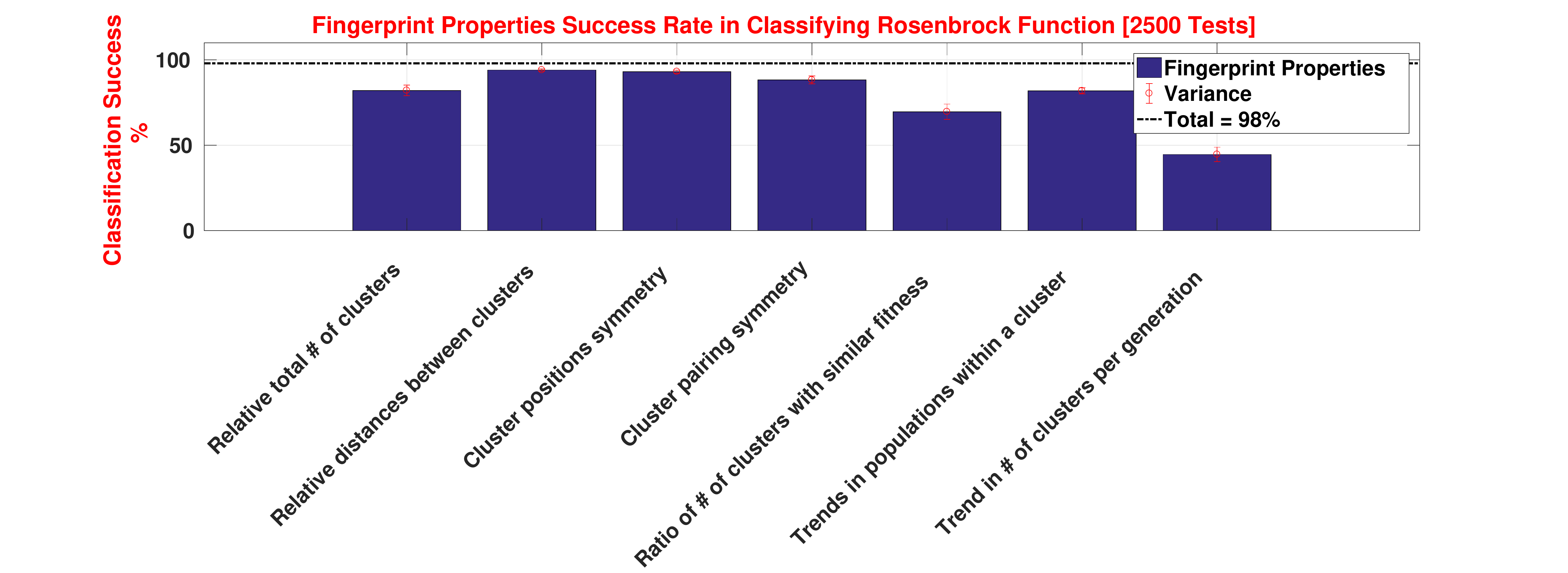}
   \caption{Classification of the Rosenbrock Function}
   \label{rosenbrockClassification}
   \vspace{-1cm}
   \end{figure}

\subsection{KIEA performance}
\label{KIEAPerformance}
    
   We conclude the experimental validation of KIEA by measuring the performance gain in terms of convergence time. Table \ref{evalTable} summarizes the convergence time on the Ackley and Gaussian function in 10 dimensions, with and without KIEA. Tests are done 25 times per function and the 1st, 7th, 13th, 19th and 25th best convergence times out of the 25 runs are captured. All runs are done with FES = $10^3$ and termination error = $10^{-6}$. It is clear that there a decrease across all the best times when using KIEA, which reaches a 80\% decrease in the worst case. Moreover, there is a significant decrease in standard deviation across all the runs, which suggests a more reliable performance when using KIEA.

\begin{table}
\caption{Convergence time T [sec] with and without KIEA}
\label{evalTable}
\centering
\begin{tabular}{|c|c|c|c|c|}
\hline
 & Ackley   & Ackley & Gaussian & Gaussian \\
 &  (w/o KIEA) &   (w/ KIEA) & (w/o KIEA) &  (w/ KIEA) \\
\hline
1st  & 1.439 & 1.261 & 0.232 & 0.1179\\
7th &  2.994  & 2.890 & 0.131 & 0.1194\\
13th & 3.014 & 2.905 & 0.137 & 0.1200\\
19th & 3.053 & 2.929 & 0.14 & 0.1283\\
25th & 10.056 & 2.990 & 0.191 & 0.1658\\
Mean & 3.677 & 2.791 & 0.1662 & 0.1303\\
Std. & 2.324 & 0.388 &  0.0439 & 0.0203\\
 \hline
\end{tabular}
\end{table}
    
    \section{Conclusions}
    \label{conclusions}
    In this work we have introduced a framework for knowledge integration in evolutionary algorithms. The framework, named KIEA, is based on the concept of \emph{EA fingerprint}, i.e., a set of a properties that capture the population behavior in the solution space while the EA is running. In addition to the algorithmic description of the framework, we presented a mathematical formalization of the properties constituting the fingerprint.
    
    In the preliminary experiments conducted in this study, the framework prototype showed a successful classification probability between 90\% and 98\%, depending on the function to optimize. Furthermore, the comparison of the convergence time on functions optimized with and without KIEA proved that the presented framework consistently enhances the convergence time, reaching a worst-case improvement of nearly 80\%. 
    
    In addition to the improved numerical performance, the presented framework has the advantage of being fully expandable, since it is possible to add new pilot functions and strategies in a straightforward manner. Furthermore, KIEA is suitable for any evolutionary algorithm as it is not strictly bound to any specific EA implementation. In future works, we will test this framework on a wider range of optimization problems, and we will expand our knowledge base of pilot functions and corresponding strategies. Finally, we plan to perform tests with different kinds of state-of-the-art EAs, to show the general applicability of KIEA.\\

 	\setlength{\tabcolsep}{0cm}
 	\noindent
     \begin{tabular}{p{0.17\linewidth} p{0.83\linewidth}}
     \raisebox{-0.8cm}{\includegraphics[height=.95cm]{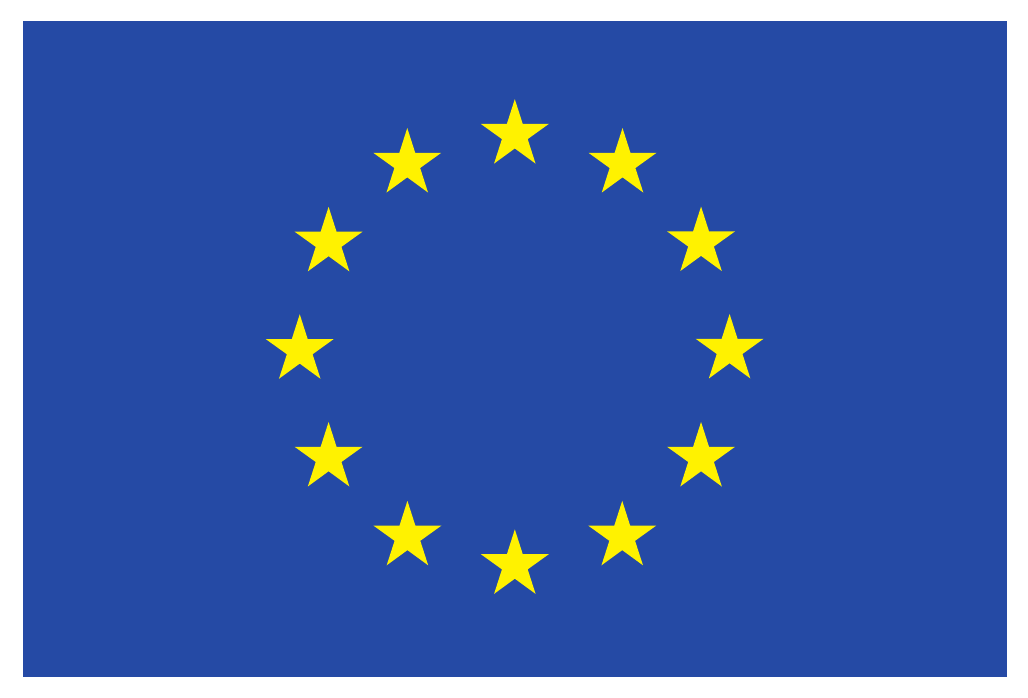}}
     &
 	{\textbf{Acknowledgments.} This project has received funding from the European Union's Horizon 2020 research and innovation program under grant agreement No 665347. We also gratefully acknowledge}
 	\end{tabular} \\
    \noindent
 the computational resources provided by RWTH Compute Cluster from RWTH Aachen University under project RWTH0118.

	\bibliographystyle{splncs03.bst}
	\bibliography{main}

\end{document}